\documentclass[10pt,twocolumn,letterpaper]{article}

\usepackage[pagenumbers]{cvpr} 

\usepackage{algorithm}
\usepackage{algorithmic}
\usepackage{multirow}
\usepackage{xcolor}
\usepackage{colortbl}
\usepackage{amssymb}
\usepackage{caption}
\usepackage{marvosym}

\definecolor{cvprblue}{rgb}{0.21,0.49,0.74}
\usepackage[pagebackref,breaklinks,colorlinks,allcolors=cvprblue]{hyperref}

\def\paperID{***} 
\def\confName{CVPR}
\def\confYear{2026}

\title{EventSTU: Event-Guided Efficient Spatio-Temporal Understanding \\for Video Large Language Models}
\author{Wenhao Xu \hspace{8pt} Xin Dong \hspace{8pt} Yue Li \hspace{8pt} Haoyuan Shi \hspace{8pt} Zhiwei Xiong\textsuperscript{\Letter}\\
University of Science and Technology of China\\
{\tt\small wh-xu@mail.ustc.edu.cn, zwxiong@ustc.edu.cn}\\
\small\url{https://wh-xu1.github.io/EventSTU.github.io/}
}

\begin{document}
\twocolumn[{
\renewcommand\twocolumn[1][]{#1}
\maketitle
\vspace{-9mm}
\begin{center}
    \includegraphics[width=1.\linewidth]{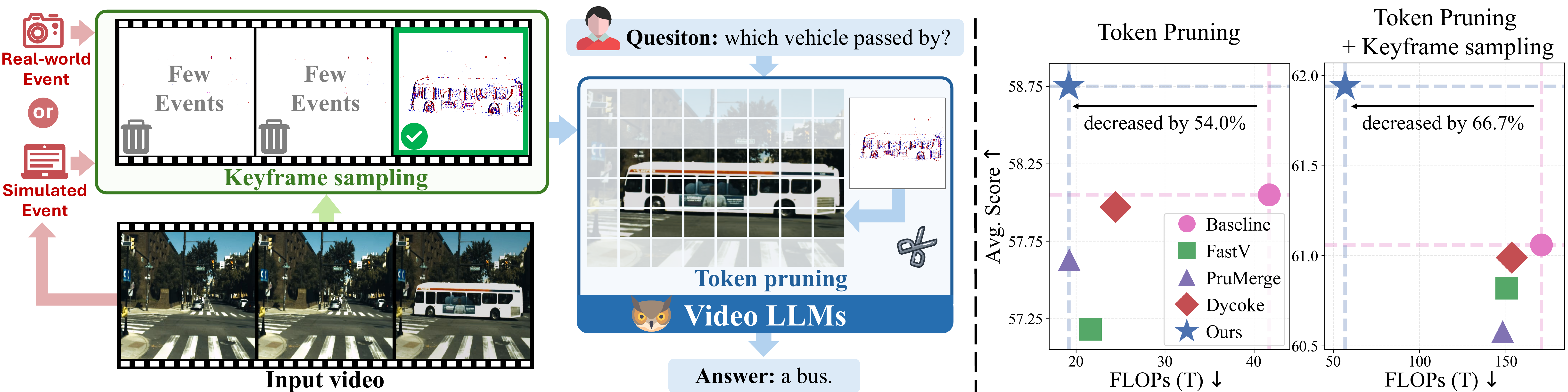}
    \vspace{-6.7mm}
    \captionsetup{type=figure}
    \caption{
    \textbf{Left}: 
    Our key innovation is the dual exploitation of events: leveraging their change-triggered property to eliminate redundant frames, and utilizing their inherent visual saliency to prune less important tokens. Notably, our framework can be driven by either real-world events or simulated events.
    \textbf{Right}:
    Our framework breaks the conventional efficiency-performance trade-off, reducing FLOPs by 66.7\% yet still outperforming the original model. This holds true whether using token pruning alone or combined with keyframe sampling.
    }
    \label{fig:teaser}
\end{center}
}]


\begin{abstract}
Video large language models have demonstrated strong video understanding capabilities but suffer from high inference costs due to the massive number of tokens in long videos. Inspired by event-based vision, we propose an event-guided, training-free framework for efficient spatio-temporal understanding, named EventSTU. In the temporal domain, we design a coarse-to-fine keyframe sampling algorithm that exploits the change-triggered property of event cameras to eliminate redundant frames. In the spatial domain, we design an adaptive token pruning algorithm that leverages the visual saliency of events as a zero-cost prior to guide spatial reduction. From a holistic spatio-temporal perspective, we further integrate question relevance from keyframe sampling to adaptively allocate token pruning budgets. To facilitate evaluation, we construct EventBench, the first event-inclusive, human-annotated multimodal benchmark that covers diverse real-world scenarios. Beyond physical event cameras, EventSTU also supports general video understanding using simulated events. Comprehensive experiments show that EventSTU achieves 3.01× FLOPs reduction and 3.10× prefilling speedup over the strongest baseline while still improving performance.
\end{abstract}

\section{Introduction}
\label{sec:intro}

Video Large Language Models (Video-LLMs) have made significant progress in complex video understanding tasks \cite{maaz2024video, zhang2023video, li2024llava, zhang2024video, li2023videochat, lin2023video, wang2024qwen2}. However, 
the massive number of frames in long videos produces tens of thousands of visual tokens. As the attention mechanism scales quadratically with token length, processing such sequences leads to prohibitive training and inference costs. 
In extreme cases, the input sequence may even exceed the maximum token capacity of LLMs, causing failure \cite{tang2025adaptive, yuframe, shenlongvu}. 
On the other hand, the redundant or irrelevant segments in long videos introduce substantial noise or uninformative tokens, which distract attention and hinder effective reasoning \cite{tao2025dycoke, liu2025bolt}
These challenges raise a critical question: \textit{how can we effectively reduce the number of visual tokens while maintaining model performance?}

In this work, we draw inspiration from biological vision systems, which achieve remarkable efficiency by prioritizing change and motion cues \cite{berry1999anticipation, gollisch2010eye, albright1995visual, gabbiani1999computation}, enabling the system to focus neural resources on critical information and process massive visual information rapidly. Event cameras \cite{gallego2020event, chakravarthi2024recent} mimic this mechanism by asynchronously capturing only brightness changes at individual pixels, producing sparse yet information-rich event streams. 
Motivated by event-based vision, we propose EventSTU, a training-free framework that leverages event guidance to eliminate both temporal and spatial redundancy, thereby facilitating efficient long video understanding.

For temporal redundancy across frames, downsampling video frames is a straightforward way. However, the uniform sampling strategy adopted by most existing approaches \cite{li2024llava, wang2024qwen2, zhang2024video} always misses critical visual cues and causes key information loss. To address this limitation, we design a \textbf{C}oarse-to-\textbf{F}ine \textbf{S}ampling algorithm, termed C2FS. At the coarse stage, C2FS exploits the change-triggered property of event cameras by using event density as a proxy for information increment. A cumulative sampling scheme is proposed to adaptively select frames rich in motion or scene variation, effectively filtering temporal redundancy. At the fine stage, C2FS further applies question-relevant sampling, selecting the most semantically relevant frames conditioned on the input question. The hierarchical sampling narrows the focus from redundancy elimination to semantic alignment, enabling efficient yet informative video understanding.


For spatial redundancy within frames, existing token pruning techniques can be broadly categorized into inner-LLM and outer-LLM approaches. Inner-LLM pruning \cite{chen2024image, yang2025topv, tao2025dycoke, dhouib2025pact, shao2025holitom} operates after early layers, yielding limited computational savings. Outer-LLM pruning \cite{shang2025llava, yang2025visionzip, alvar2025divprune, sun2025llava} computes pairwise token similarities, resulting in quadratic overhead. To address these issues, we design a \textbf{Z}ero-cost \textbf{A}daptive \textbf{P}runing algorithm, termed ZAP, which mimics the human cognitive process from physical perception to semantic focus. Specifically, ZAP first exploits event sparsity as a natural saliency cue for physics-aware pruning, and subsequently reuses vision encoder attention scores for semantic-aware pruning. Critically, rather than pruning each frame in isolation, ZAP integrates question relevance derived from C2FS to adaptively allocate per-frame pruning budgets. This joint design enables a more informed and balanced token allocation with a holistic spatio-temporal view.

To rigorously evaluate event-guided video reasoning, we construct EventBench, the first event-inclusive and human-annotated multimodal benchmark. EventBench provides high-quality event streams and RGB videos across diverse real-world scenarios, including 8 task types with different focuses. Multiple expert annotators carefully design and annotate 500 multiple-choice question-answer pairs that probe temporal localization, spatial grounding, and holistic video understanding. 


Interestingly, EventSTU also generalizes beyond real-world events to conventional videos through simulated event guidance. Given that existing Video-LLMs \cite{li2024llava, zhang2024video, wang2024qwen2} typically process low frame-rate inputs, fine-grained microsecond event details are unnecessary. Thus, we utilize a lightweight, on-the-fly event simulation technique \cite{lou2025v2v} that directly generates event frames \cite{weng2022boosting} from adjacent image frames, offering low-latency event guidance.

Our main contributions can be summarized as follows:
\begin{itemize}

\item We propose EventSTU, an event-guided framework for efficient spatio-temporal understanding, which comprises two elaborate strategies: C2FS to retrieve non-redundant, question-relevant keyframes, and ZAP to distill visually salient and semantically rich tokens
\item We construct EventBench, the first human-annotated multimodal benchmark that incorporates real event streams for evaluating event-assisted reasoning capability.
\item We extend EventSTU to conventional video understanding with simulated events, generalizing the event-inspired efficiency beyond real event data.
\item Experimental results on four benchmarks show that EventSTU significantly reduces the number of tokens while improving performance with both real and simulated events.
\end{itemize}
\section{Related Work}
\label{sec:related}

\begin{figure*}[t!]
  \centering
  \includegraphics[width=1.\textwidth]{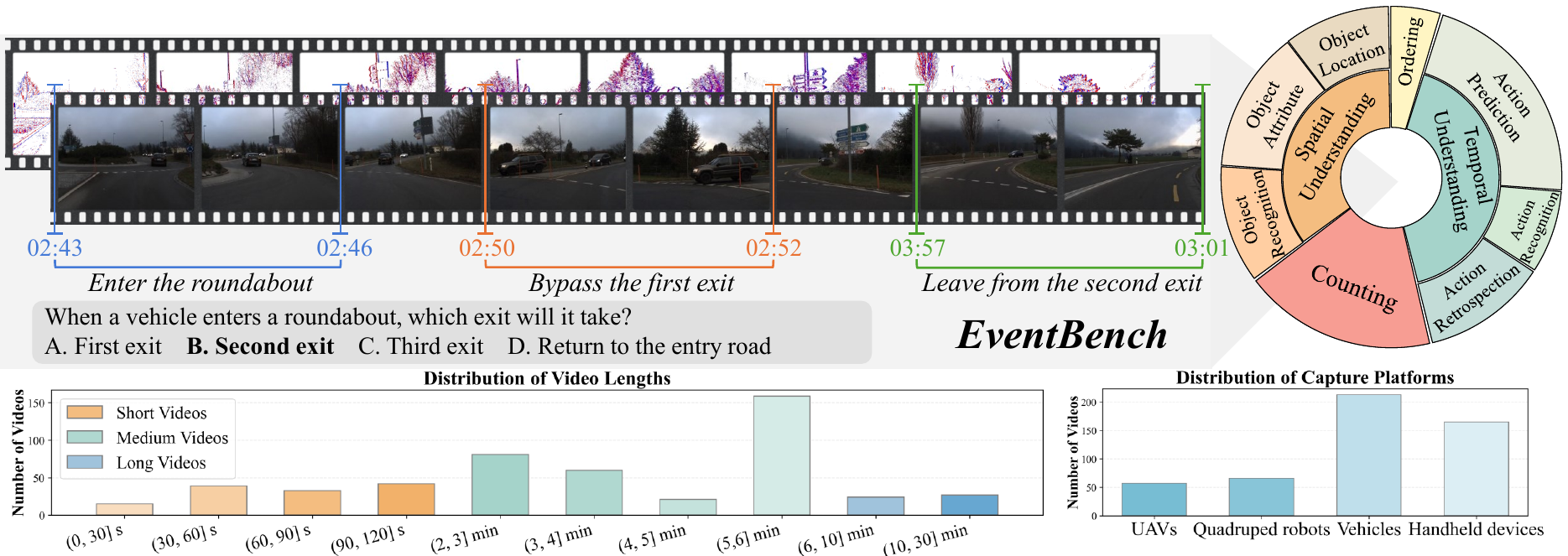}
  \vspace{-7mm}
  \caption{\textbf{Top}: A sunburt and toy example of the proposed EventBench. \textbf{Bottom}: Statistics of EventBench including the distributions of video lengths and capture platforms.}
  \label{fig:benchmark}
  \vspace{-5mm}
\end{figure*}

\subsection{Token Reduction in Video-LLMs}

\noindent\textbf{Keyframe sampling.} The performance of video-LLMs depends on the quality of the sampled frames. VideoAgent \cite{wang2024videoagent} and VideoTree \cite{wang2025videotree} generate frame captions and utilize LLM to select keyframes, while \cite{hu2025m} trains a multimodal LLM as a specialized keyframe selector. BOLT \cite{liu2025bolt} employs vision-language models to select question-relevant frames, and AKS \cite{tang2025adaptive} focuses on balancing question relevance with sampling diversity. However, they rely on expensive, heavy-model pre-processing. Instead, we leverage event density to efficiently eliminate redundant frames, reducing the substantial computational cost and achieving a superior performance–efficiency trade-off.

\noindent\textbf{Token pruning.} Visual token pruning has gained increasing attention. FastV \cite{chen2024image} uses attention scores to prune tokens after the early LLM layers, while PACT \cite{dhouib2025pact} further proposes alternative importance scores to enable FlashAttention \cite{dao2022flashattention} compatibility. DyCoKe \cite{tao2025dycoke} introduces dynamic key-value cache pruning during the decoding stage. However, these methods yield limited FLOPs reduction as they leave early layers unpruned. LLaVA-PruMerge \cite{shang2025llava} and VisionZip \cite{yang2025visionzip} merge tokens before the LLM based on pairwise token similarities, incurring quadratic computational overhead. In contrast, we leverage the zero-cost visual saliency of events to prune tokens before LLMs.

\subsection{Event-based Understanding}
Unlike mature image understanding, the event community is still in the early stages of developing universal models. EventCLIP \cite{wu2023eventclip} first adapts CLIP to align events with language for zero-shot event recognition. CEIA \cite{xu2024ceia} and EventBind \cite{zhou2024eventbind} center on images to learn unified representations, mitigating event-text data scarcity. OpenESS \cite{kong2024openess} extends this alignment to open-vocabulary event segmentation. To handle more complex interactive tasks, EventGPT \cite{liu2025eventgpt} and EventVL \cite{li2025eventvl} develop event-based LLMs, but they are limited to short event streams ($<$100ms). These works share a common motivation: leveraging the robustness of events to address reasoning tasks under challenging imaging conditions. Differently, we focus on the efficiency of events, opening a new avenue for event-based LLMs.

\section{EventBench Construction}
\label{sec:eventbench}

We construct EventBench, the first event-inclusive, human-annotated multimodal benchmark, crafted for evaluating video-LLMs on event-boosted long video understanding. The statistics of our benchmark are shown in \cref{fig:benchmark}.

\noindent\textbf{Data collection.} We curate a diverse collection of high-quality, high-resolution event streams \cite{chaney2023m3ed, gehrig2021dsec, kim2024towards, liu2024seeing}, each with spatially aligned and temporally synchronized RGB videos. To cover the application scenarios of event cameras, our benchmark encompasses four capture platforms: UAVs, ground vehicles, quadruped robots, and handheld devices, across three environmental conditions: indoor, outdoor daytime, and outdoor nighttime. Furthermore, we filter out many short sequences, resulting in event streams averaging 4.3 minutes in duration, with the longest reaching 23.5 minutes. This positions EventBench as a valuable benchmark for long video understanding.

\noindent\textbf{Question design.} We follow two fundamental design principles from existing multimodal benchmarks \cite{fu2025video, wu2024longvideobench, zhou2024mlvu, huang2024vbench, wang2025lvbench}: human annotation and multiple-choice format. These ensure evaluation reliability and quantifiability. Specifically, building on the collected data, multiple annotators meticulously design and rigorously validate 500 multiple-choice questions, covering 8 diverse tasks: object recognition, object attribute recognition, object localization, action recognition, action prediction, action retrospection, counting, and ordering. Among these, counting and ordering assess holistic video understanding, while the remaining tasks assess the localization and contextualization of specific segments. We believe that EventBench will advance the research of event-assisted LLMs by providing a comprehensive and reliable evaluation.

\begin{figure*}[t!]
  \centering
  \includegraphics[width=1.\textwidth]{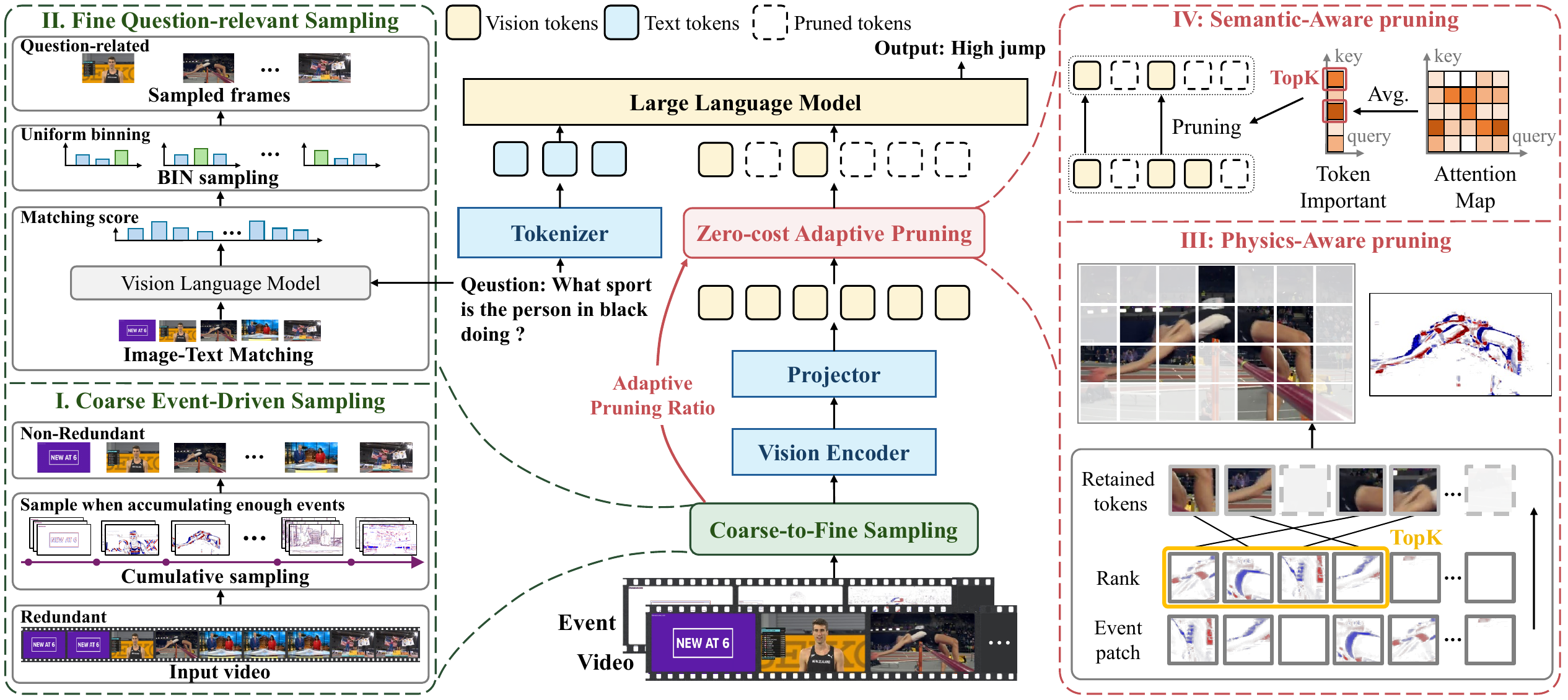}
  \vspace{-7mm}
  \caption{\textbf{Overview of EventSTU.} It processes long videos through a sequential, multi-stage pipeline. First, coarse-to-fine sampling efficiently filters redundant frames based on event density and retrieves question-relevant keyframes. Subsequently, physics-aware pruning selects tokens with high event saliency, while semantic-aware pruning further distills them to the most semantically crucial tokens using attention scores. This entire process culminates in a compact yet semantically dense visual representation, tailored for LLM inference.}
  \label{fig:main}
  \vspace{-4mm}
\end{figure*}

\section{Event-Guided Video Understanding}
\label{sec:method}

\subsection{Overview}
To address the challenges in Video-LLMs, we propose a training-free framework (EventSTU) that achieves efficient video reasoning by coupling temporal keyframe sampling and spatial token pruning under a unified event-guided paradigm. An overview is illustrated in \cref{fig:main}.

Given an RGB video sequence $\{F_{t}\}_{t=1}^{M}$, we additionally incorporate a sequence of event frames $\{E_{t}\}_{t=1}^{M-1}$, where each $E_t$ encodes the accumulated brightness changes between consecutive frames $F_{t-1}$ and $F_t$. These event frames can either originate from real sensors or be synthesized via simulators.

EventSTU comprises two components: coarse-to-fine sampling (C2FS) and zero-cost adaptive pruning (ZAP) algorithms. In the first step, C2FS reduces the video from $M$ original frames to $M_c$ non-redundant frames, and then selects the $M_f$ most question-relevant keyframes, where $M_f \ll M_{c} \ll M$. After tokenization and projection of the selected keyframes, ZAP sequentially applies physics-aware pruning and semantic-aware pruning with adaptive pruning ratios to filter out less important tokens. This yields compact yet semantically rich visual tokens.

Finally, the visual tokens and textual tokens derived from the question are concatenated and fed into the LLMs. Through the synergy of C2FS and ZAP, EventSTU delivers both computational efficiency and strong video reasoning performance, all without training or fine-tuning.

\begin{figure}[t]
\vspace{-3mm}
\begin{algorithm}[H]
\caption{Cumulative Sampling}
\label{alg:cumulative_sampling}

\textbf{Input:} Event density $\{e_t\}_{t=1}^{M-1}$, sampling rate $S \in [0,1]$

\textbf{Output:} Sampled indices $\mathcal{I}$

\begin{algorithmic}[1]
\setlength{\algorithmicindent}{2em}
\STATE Compute threshold: $\tau = \frac{\sum_{t=1}^{M-1} e_t}{\lceil (M-1)S \rceil}$
\STATE Initialize: accumulator $a = 0$, indices $\mathcal{I} = \emptyset$

\FOR{$t = 1$ to $M-1$}
\STATE $a = a + e_t$
\IF{$a \geq \tau$}
\STATE $\mathcal{I} = \mathcal{I} \cup \{t\}$, $a = 0$
\ENDIF
\ENDFOR

\RETURN $\mathcal{I}$ 
\end{algorithmic} 
\end{algorithm}
\vspace{-8mm}
\end{figure}

\subsection{Coarse-to-fine Sampling}
\label{sec:keyframe_sampling}

In long video understanding tasks, an effective keyframe sampling strategy must address two critical challenges: minimizing temporal redundancy and ensuring question relevance.
Build on the principles, we deisgn a coarse-to-fine sampling (C2FS) algorithm that progressively refines sampling from temporal redundancy removal to question relevance retrieval.


\noindent\textbf{Coarse Event-driven Sampling Stage.} 
The change-triggered property of event data makes it an ideal candidate for efficiently identifying frames with significant motion or scene variations, thus serving as a proxy for information increment. Given the input video, a naive approach of selecting frames based solely on event density could result in over-sampling in high-density regions and under-sampling in potentially important low-density areas.

To address this, we propose a cumulative sampling (CS) algorithm, which adaptively selects frames in a balanced manner based on accumulated density. Specifically, the algorithm first computes the event density sequence $\{e_t\}_{t=1}^{M-1}$, where each $e_t$ quantifies the amount of change in frame $F_t$ relative to $F_{t-1}$. For a given sampling ratio $S$, a threshold $\tau$ is computed to determine when to sample a frame. The algorithm accumulates event density and triggers a sampling action when the accumulated density exceeds $\tau$. This process repeats until the entire sequence has been processed, producing a set of keyframes that efficiently removes temporal redundancy. The detailed CS algorithm is presented in \cref{alg:cumulative_sampling}.


\noindent\textbf{Fine Question-relevant Sampling Stage.}
\label{sec:fine_sampling}
To further retrieve keyframes, we incorporate question-relevance by leveraging the text-image similarity. In this stage, we first encode the question and coarse-sampled frames 
by the textual and visual encoders respectively, such as BLIP \cite{li2022blip} or CLIP \cite{radford2021learning}. The separate embeddings for question and frames are then computed the cosine similarity scores.

Instead of directly selecting $M_{f}$ frames with top scores, we introduce a bin sampling strategy to ensure the selection spans a broad temporal range. Specifically, we uniformly divide the coarse-sampled frames into $B$ bins based on their temporal position and select the frame with the highest similarity score from each bin.
This method ensures that the selected frames maintain temporal diversity while simultaneously being highly relevant to the user question.

Put things together, C2FS decouples the keyframe sampling into redundancy removal and relevance retrieval. 
The coarse event-driven sampling serves as an efficient, content-agnostic filter, eliminating redundant frames and reducing the computational burden for subsequent processing. The fine question-relevant sampling ensures the selected frames are temporally diverse and highly relevant to the question.



\subsection{Zero-cost Adaptive Pruning}
\label{sec:token_pruning}
Even after reducing temporal redundancy across the keyframes, spatial redundancy within the selected frames remains a significant challenge. To address this limitation, we design a Zero-cost Adaptive Pruning (ZAP) algorithm. ZAP operates in two phases: physics-aware pruning and semantic-aware pruning. These phases are applied sequentially to the set of visual tokens derived from the selected keyframes. Besides, ZAP adaptively allocates the pruning ratio for each frame by integrating the question relevance from C2FS, thereby achieving holistic token pruning.

\begin{figure}[t!]
  \centering
  \includegraphics[width=1.0\linewidth]{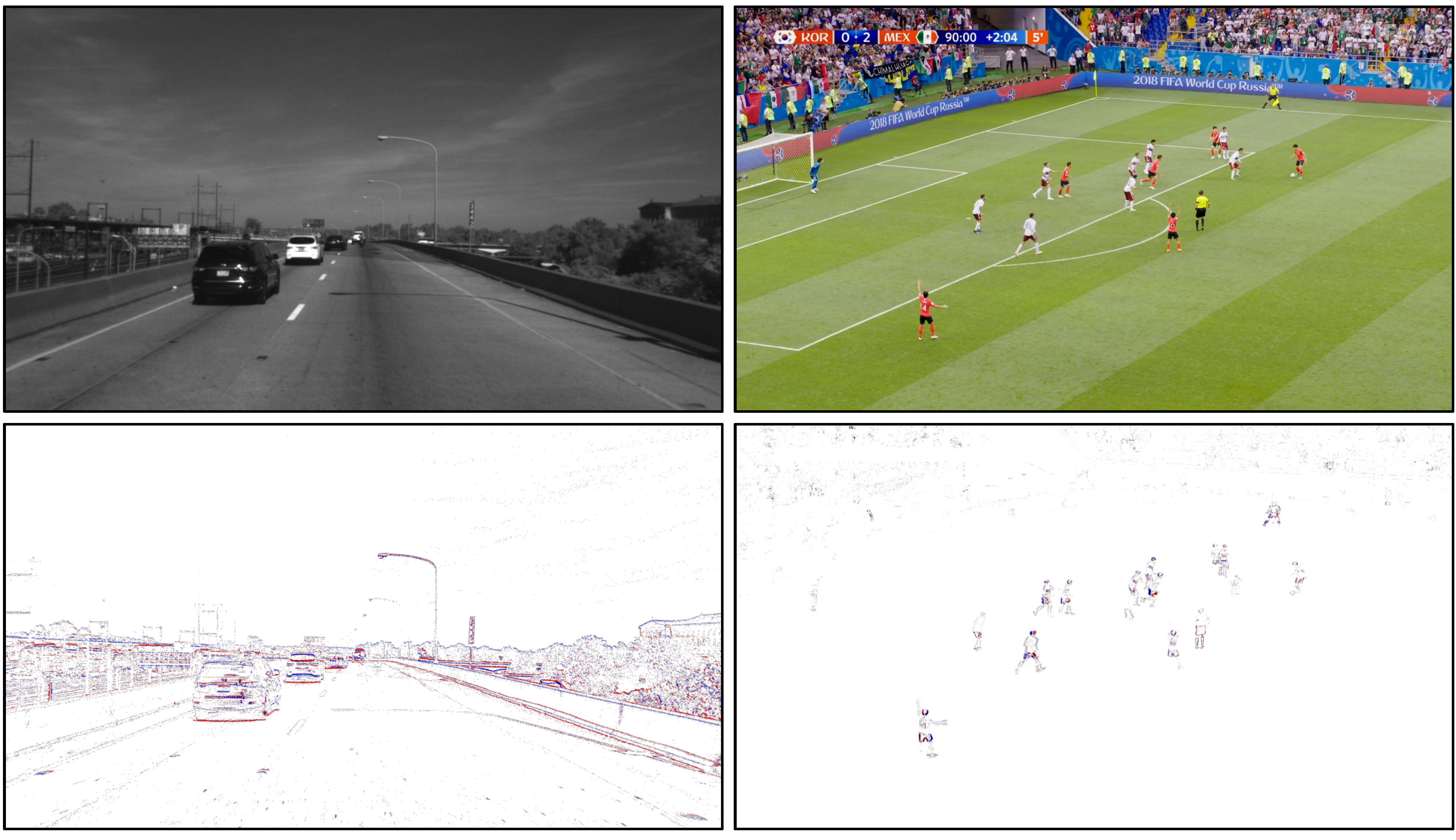}
  \vspace{-6mm}
  \caption{\textbf{Visual saliency of event data.} \textbf{Left:} The large, uninformative sky and roads trigger few events. \textbf{Right:} Running athletes that attract more user interest trigger significantly more events.
  }
  \label{fig:sample}
  \vspace{-3mm}
\end{figure}

\noindent\textbf{Physics-aware pruning.}
Event data inherently provides more compact representations due to its sparse encoding of brightness changes. As shown in \cref{fig:sample}, 
this sparsity manifests in two ways: it filters out uninformative regions, such as the sky and roads, which trigger few events but occupy large portions of visual tokens. At the same time, it highlights changing regions that trigger more events and typically attract greater user interest. Building upon these insights, we introduce event-based physics-aware pruning.

Specifically, we partition an event frame into non-overlapping patches, each corresponding to an RGB visual token. We then compute the event density within each patch, using it as an importance score. For each RGB frame, we prune the $K_{p}$\% of tokens with the lowest scores, effectively filtering out the uninformative regions. Our physics-aware pruning is simple yet highly effective, with nearly zero computational cost.

\noindent\textbf{Semantic-aware pruning.}
In addition to physical pruning, we introduce semantic-aware pruning, which leverages the attention scores of the vision encoder to identify semantically significant tokens. Specifically, we reuse the attention matrix from the last layer of the vision encoder to assess the importance of each token. By averaging the attention scores across the query dimension, we obtain a per-token importance measure. Visual tokens with higher average attention are deemed more relevant, while tokens with lower attention scores are considered less significant.

We prune the $K_{s}$\% of tokens that receive the least attention, further refining the set of tokens retained for subsequent processing. This approach benefits from the efficient reuse of the attention matrix, ensuring negligible computational overhead.

\noindent\textbf{Adaptive Pruning Ratio.}
To achieve holistic token pruning across keyframes, we introduce a strategy that adaptively allocates pruning ratios based on their relevance to the input question. The goal is to ensure that the more relevant keyframes retain more visual tokens, while less relevant keyframes undergo more aggressive pruning.

Mathematically, the total allocable token budgets $N_{\text{total}}$ is $M_f \cdot N \cdot (1 - K - R)$, where $M_f$ is the number of keyframes, $N$ is the number of tokens per frame, $K$ is the expected overall pruning ratio, and $R$ is the base retained ratio (empirically set to 5\% to avoid over-pruning, with $K<1-R$). The adaptive pruning ratio $K^t$ for each frame is defined as:
\begin{equation}
K^t = \frac{N_{\text{total}} \cdot s_t + R \cdot N}{N},
\end{equation}
where $s_t$ is the normalized similarity score to the question, as derived from the C2FS. 

To balance the physics-aware pruning ratio $K_p$ and semantic-aware pruning ratio $K_s$, we set:
\begin{equation}
(K_p^t, K_s^t) = 
\begin{cases}
(K^t, 0) & \text{if } K^t \leq K_p, \\
(K_p, 1 - \frac{1-K^t}{1-K_p}) & \text{otherwise},
\end{cases}
\end{equation}
where $K_p^t$ and $K_s^t$ denote the ratios
in physics-aware and semantic-aware pruning for frame $F_t$, respectively.




\begin{table*}[t]
    \centering
    \caption{\textbf{Comparison on three settings across four benchmarks:} EventBench (EB), Video-MME (V-MME), LongVideoBench (LVB), and MLVU. Here, ``Token Budget'' denotes the retained token ratio after pruning. ``ZAP\textsubscript{phy}'' denotes using physics-aware pruning only.}
    \vspace{-3mm}
    \setlength{\tabcolsep}{6pt}
    \renewcommand{\arraystretch}{1}
    \resizebox{1.0\linewidth}{!}{
        \begin{tabular}{c|c|cc|cccc|c}
        \toprule

    
        
        Method & Token Budget & FLOPs (T)$\downarrow$ & FLOPs Ratio$\downarrow$ & EB$\uparrow$ & V-MME$\uparrow$ & LVB$\uparrow$ & MLVU$\uparrow$ & Avg.$\uparrow$ \\
        
        \midrule
        \rowcolor{gray!15}
        LLaVA-OV-7B \cite{li2024llava} & 100\% & 41.71 & 100\% & 52.50 & 58.48 & 56.40 & 64.81 & 58.05 \\  

        \midrule
        \multicolumn{9}{c}{\textit{Keyframe Sampling}} \\
        \midrule
        
        BOLT \cite{liu2025bolt} & 100\% & 170.6 & 409\% & - & 59.90 & 59.60 & 66.80 & - \\
        AKS \cite{tang2025adaptive} & 100\% & 170.6 & 409\% & 54.09 & 60.44 & 60.28 & 69.41 & 61.06 \\
        \textbf{C2FS (Ours)} & 100\% & \textbf{79.34} & \textbf{190}\% & \textbf{55.09} & \textbf{60.96} & \textbf{60.43} & \textbf{69.87} & \textbf{61.59} \\

        \midrule
        \multicolumn{9}{c}{\textit{Token Pruning}} \\
        \midrule

        FastV \cite{chen2024image} & 50\% & 21.61 & 51.8\% & 50.90 & 57.74 & 56.10 & 63.98 & 57.18 \\
        PruMerge \cite{shang2025llava} & 50\% & 19.24 & 46.1\% & 51.90 & 58.04 & 55.95 & 64.63 & 57.63 \\
        DyCoke \cite{tao2025dycoke} & 50\% & 24.45 & 58.6\% & 52.50 & \textbf{59.04} & 55.80 & 64.54 & 57.97 \\
        \textbf{ZAP\textsubscript{phy} (Ours)} & 50\% & \textbf{19.20} & \textbf{46.0}\% & 52.89 & 58.63 & 56.62 & 64.63 & 58.19 \\
        \textbf{ZAP (Ours)} & 50\% & \textbf{19.20} & \textbf{46.0}\% & \textbf{53.89} & 58.85 & \textbf{56.99} & \textbf{65.27} & \textbf{58.75} \\

        \midrule
        \multicolumn{9}{c}{\textit{Keyframe Sampling + Token Pruning}} \\
        \midrule

        AKS \cite{tang2025adaptive} + FastV \cite{chen2024image} & 50\% & 150.5 & 361\% & 53.49 & 60.22 & 60.28 & 69.27 & 60.82 \\
        AKS \cite{tang2025adaptive} + PruMerge \cite{shang2025llava} & 50\% & 148.1 & 355\% & 52.10 & 60.48 & 59.84 & 69.92 & 60.58 \\
        AKS \cite{tang2025adaptive} + DyCoke \cite{tao2025dycoke} & 50\% & 153.3 & 368\% & 53.49 & 60.56 & 59.91 & 70.01 & 60.99 \\
        \rowcolor{blue!8}
        \textbf{EventSTU (Ours)} & 50\% & \textbf{56.83} & \textbf{136}\% & \textbf{55.49} & \textbf{61.04} & \textbf{60.66} & \textbf{70.56} & \textbf{61.94} \\

        \midrule
        AKS \cite{tang2025adaptive} + FastV \cite{chen2024image} & 30\% & 143.4 & 344\% & 51.10 & 58.89 & 58.56 & 69.41 & 59.49 \\
        AKS \cite{tang2025adaptive} + PruMerge \cite{shang2025llava} & 30\% & 140.2 & 336\% & 52.89 & 59.89 & 57.59 & 69.69 & 60.02 \\
        AKS \cite{tang2025adaptive} + DyCoke \cite{tao2025dycoke} & 30\% & 147.1 & 353\% & 53.89 & 60.41 & 59.16 & 70.61 & 61.02 \\
        \rowcolor{blue!8}
        \textbf{EventSTU (Ours)} & 30\% & \textbf{48.93} & \textbf{117}\% & \textbf{54.89} & \textbf{61.44} & \textbf{59.99} & \textbf{70.75} & \textbf{61.77} \\
        
        \bottomrule
      \end{tabular}
    }
    \label{tab:main}
    \vspace{-3mm}
\end{table*}

\section{Extension to General Video Understanding}
\label{sec:event_simulation}
Event-based data is known for its superior efficiency, prompting the question: \textit{Can our framework be driven by simulated events instead of relying on physical event cameras?} Traditional event simulation techniques \cite{hu2021v2e, gehrig2020video, ziegler2023real} typically interpolate video to extremely high frame rates, introducing significant computational cost. To avoid this, we utilize a lightweight, on-the-fly event simulation method \cite{lou2025v2v} tailored to general video understanding, leveraging frame differences to simulate event frames without complex interpolation.

We simulate events by calculating intensity differences between consecutive RGB frames, $F_t$ and $F_{t-1}$. After applying reverse gamma correction $\gamma^{-1}$, we compute the intensity change $\bigtriangleup L_{t, t-1}$ as:
\begin{equation}
\bigtriangleup L_{t, t-1} = \log(\gamma^{-1}(F_t)) - \log(\gamma^{-1}(F_{t-1})).
\end{equation}
Next, we calculate the number of positive and negative events at each pixel using predefined thresholds $C_p$ and $C_n$:
\begin{equation}
N_p=\left \lfloor \frac{\left [ \bigtriangleup L_{t, t-1} \right ]_+}{C_p}  \right \rfloor ,N_n=\left \lfloor \frac{\left [ -\bigtriangleup L_{t, t-1} \right ]_+}{C_n}  \right \rfloor,
\end{equation}
where $\left [ \cdot  \right ] _+$ denotes $max(0,\cdot )$.
These events are then used to construct a 2D event frame with the total event count $N_p + N_n$ at each pixel. Importantly, this method eliminates the need for microsecond-level, high-frame-rate interpolation, which is computationally expensive and unnecessary, while still capturing the second-level changes needed for video reasoning.

This approach allows us to break free from the dependence on physical event sensors, enabling EventSTU to extend efficiently to general video understanding tasks.

\section{Experiment}
\label{sec:experiment}

\subsection{Experimental Settings}

\begin{figure*}[t!]
  \centering
  \includegraphics[width=1.0\textwidth]{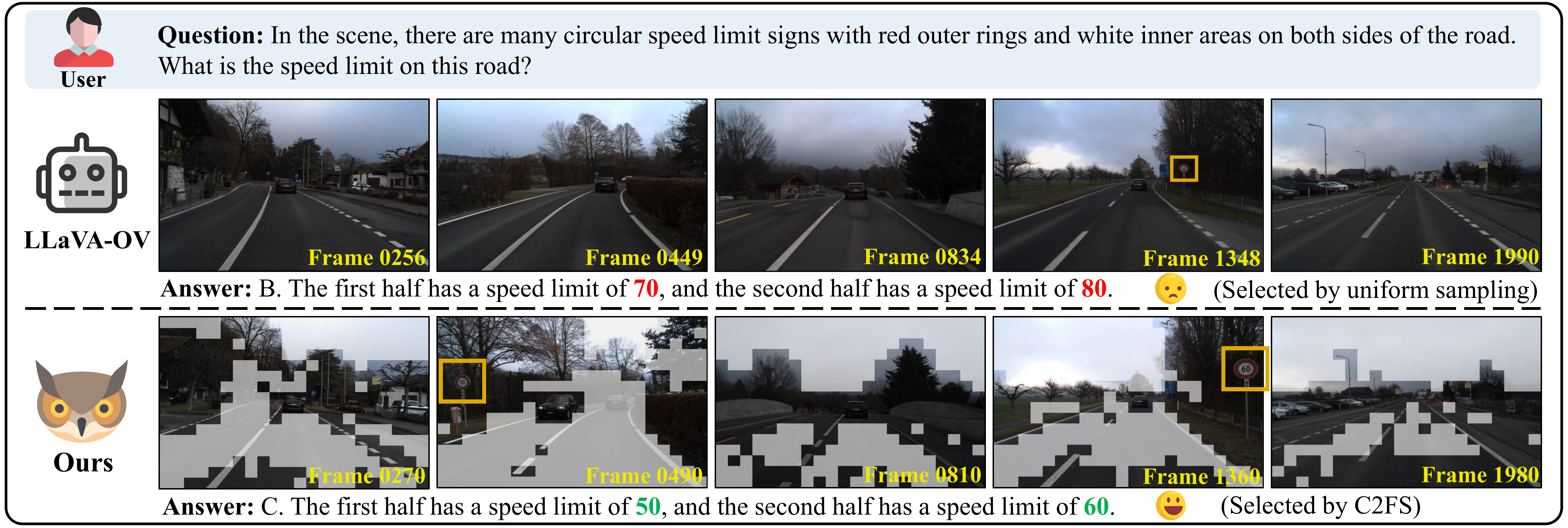}
  \vspace{-5mm}
  \caption{\textbf{Visualization results on EventBench.} ``LLaVA-OV'' represents the original model without our method. It uses uniform sampling and misses a keyframe. In contrast, our method captures all keyframes and prunes uninformative areas, highlighting the speed limit signs.}
  \label{fig:visualization}
  \vspace{-3mm}
\end{figure*}

\noindent\textbf{Benchmarks.}
We evaluate our method on our EventBench with real event data and three widely-used general video benchmarks with simulated event data: LongVideoBench \cite{wu2024longvideobench}, Video-MME \cite{fu2025video}, and MLVU \cite{zhou2024mlvu}. These benchmarks provide a comprehensive testbed for assessing the effectiveness and generalization of our method with and without physical event cameras.

\noindent\textbf{Comparison methods.}
Our method is compared in three settings: keyframe sampling, token pruning, and their combination. For keyframe sampling, we compare our C2SF against two latest training-free methods, AKS \cite{tang2025adaptive} and BOLT \cite{liu2025bolt}. Since BOLT is not publicly available, we compare against its reported results. For token pruning, we compare our ZAP against three strong training-free methods: FastV \cite{chen2024image}, LLaVA-PruMerge \cite{shang2025llava}, and DyCoKe \cite{tao2025dycoke}. For the combined setting, we compare our complete framework against AKS paired with each token pruning method.

\noindent\textbf{Implementation details.}
We implement our methods on LLaVA-OneVision-7B \cite{li2024llava} using 8 NVIDIA RTX 3090 GPUs. For video input, we follow the official settings of LLaVA-OneVision, using 32 input frames (i.e., our fine-sampling target $M_f$) and $N$ = 196 tokens per frame. We set the coarse frame sampling ratio $S$ to 25\%. The overall pruning ratio $K$ serves as the main experimental variable, with the physics-aware pruning ratio $K_p$ set to 20\% on EventBench and 25\% on general benchmarks.

\subsection{Comparison with Existing Methods}

\noindent\textbf{Keyframe sampling.}
As shown in \cref{tab:main}, while BOLT and AKS improve performance by sampling more useful keyframes but introduce a substantial 409\% increase in FLOPs. In contrast, our C2FS leverages event data to filter out redundant frames, significantly reducing computational cost. Compared to AKS, our C2FS requires 46.5\% of its FLOPs while still surpassing it by 0.53 in average performance, making keyframe sampling more practical.

\noindent\textbf{Token Pruning.}
Notably, even when employing physics-aware pruning only (i.e., ZAP\textsubscript{phy} in \cref{tab:main}), our method achieves superior performance to the original model at just 46.0\% of its FLOPs. This result validates our premise that the visual saliency of event data effectively eliminates noisy signals, leading to improved reasoning in LLMs.

With semantic-aware pruning, our full ZAP delivers stronger performance. Against FastV, ZAP surpasses FastV by 1.57 points with lower FLOPs, since FastV cannot prune early LLM layers due to its inner-LLM pruning design. While LLaVA-PruMerge has comparable FLOPs, it introduces significant inference latency (detailed in \cref{sec:latency_analysis}). In summary, at the same token budget, our ZAP simultaneously achieves the best average performance, lowest FLOPs, and minimal inference latency.

\noindent\textbf{Keyframe sampling + Token Pruning.}
While pairing AKS with token pruning methods reduces FLOPs, it incurs a performance penalty, highlighting the limitations of applying these methods independently. In contrast, our EventSTU integrates C2FS and ZAP to facilitate a more informed token allocation from a holistic spatio-temporal perspective, thereby breaking the conventional efficiency-performance trade-off. This holistic design translates into superior results: at a 50\% token budget, EventSTU outperforms ``AKS + DyCoke'' by 0.95 points while requiring only 37.1\% of its FLOPs.

Particularly, at a 30\% token budget, EventSTU significantly boosts average performance by 3.72 points over the original model with a comparable computational cost. This advantage is evident in \cref{fig:visualization}, where EventSTU helps the LLM focus on critical visual content by capturing relevant frames and pruning distracting areas. By fully exploiting the efficiency of event data, we achieve an unprecedented efficiency-performance balance, elevating keyframe sampling and token pruning to a new level of practical applicability.

\subsection{Latency Analysis}
\label{sec:latency_analysis}

\begin{figure}[t]
  \centering
  \includegraphics[width=1.0\linewidth]{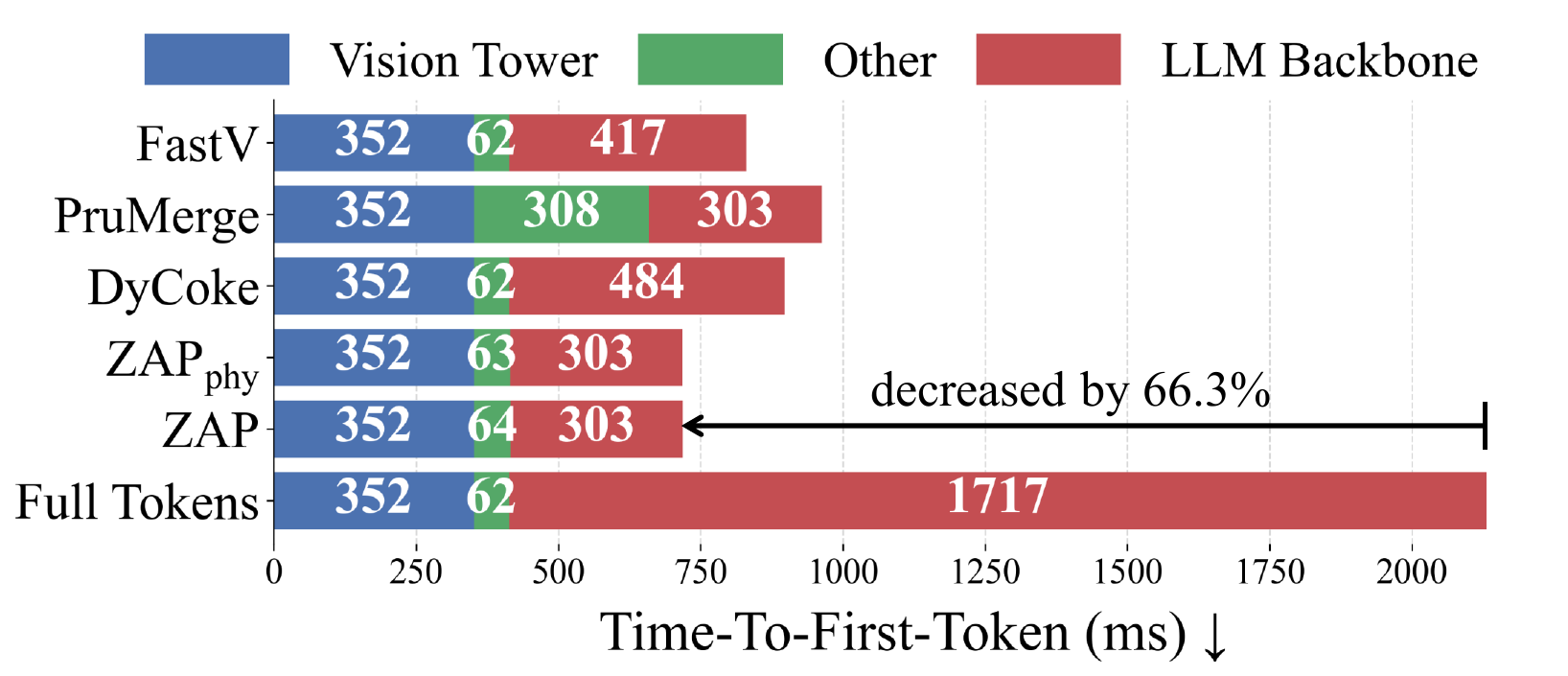}
  \vspace{-5mm}
  \caption{\textbf{Latency Analysis.} ``Other'' indicates token pre-processing time. Our ZAP achieves a 66.3\% reduction in Time-To-First-Token (TTFT) compared to the original model, outperforming all other token pruning methods.
  }
  \label{fig:TTFT}
  \vspace{-2mm}
\end{figure}

\begin{figure*}[t!]
  \centering
  \vspace{-6mm}
  \includegraphics[width=1.0\linewidth]{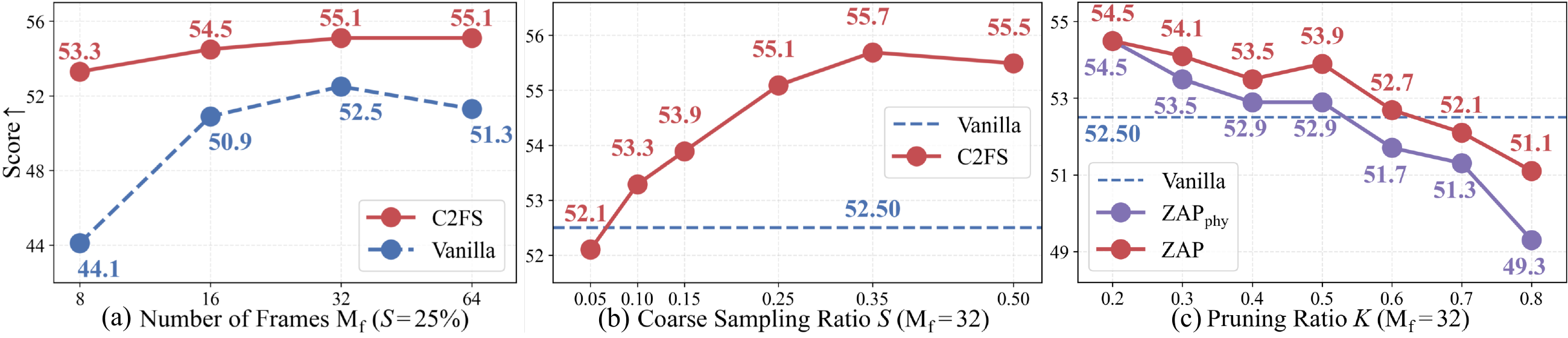}
  \caption{\textbf{Analysis of number of frames (a), coarse sampling ratio (b) in C2FS, and pruning ratio (c) in ZAP.} ``Vanilla'' denotes the original model. ``ZAP\textsubscript{phy}'' denotes using physics-aware pruning only.
  }
  \label{fig:variation}
  \vspace{-1mm}
\end{figure*}

\cref{fig:TTFT} presents a latency comparison between our ZAP and other token-pruning methods. Since our focus is on the pre-filling stage, we measure the time-to-first-token (TTFT). Specifically, LLaVA-PruMerge relies on an expensive computation of pairwise token similarities, which leads to significant pre-processing time (i.e., ``other'' time). Meanwhile, inner-LLM methods such as FastV and DyCoKe are constrained by their designs, failing to reduce latency in the early layers of the LLM. In contrast, our ZAP avoids these pitfalls. Leveraging zero-cost guidance from event data and reused attention for outer-LLM pruning, our ZAP introduces an additional pre-processing time of just 2 milliseconds. Compared to the original model, our ZAP reduces TTFT by 66.3\%---the lowest across all compared methods.

\subsection{Ablation Study}
\label{sec:ablation_study}

\noindent\textbf{Comparison with alternative sampling strategies.} 
As shown in \cref{tab:CS}, we compare alternative strategies for coarse and fine sampling on EventBench and LongVideoBench. At the coarse sampling stage, “UNI” (uniform sampling) performs worst because it fails to remove redundant frames, resulting in a lower signal-to-noise ratio. “TOP” (selecting frames with the highest event density) also struggles, as it over-samples in high-density areas and under-samples in potentially important low-density areas. In contrast, ``CS'' (cumulative sampling) adaptively sampling in a balanced manner based on accumulated density, achieving the best performance.

At the fine sampling stage, ``TOP'' (sampling frames with the highest image-text similarity) spans a narrow temporal window. Despite enabling localization of the queried segments, it fails to provide sufficient context to answer the question. In contrast, ``BIN'' (bin sampling) effectively balances question relevance and temporal coverage, significantly outperforming ``TOP'' by 1.2 points on EventBench.

\noindent\textbf{Ablation study on ZAP.} 
As shown in \cref{tab:ablation_ZAP}, combining physics-aware and semantic-aware pruning yields superior performance to applying either method in isolation. This is because the two stages form a coherent pipeline that transitions from capturing raw visual saliency to preserving deep semantic meaning. Additionally, we observe that adaptive pruning ratio improves average performance by an impressive 1.51 points. This demonstrates the effectiveness of optimizing token pruning budgets from a holistic spatio-temporal perspective.

\noindent\textbf{Analysis on number of frames.} 
\cref{fig:variation} (a) demonstrates the consistent superiority of C2FS over the original model. The improvement is particularly evident with a few frames ($M_f=8$), confirming that our C2FS effectively eliminates redundant or irrelevant frames, thereby enhancing the signal-to-noise ratio.

\noindent\textbf{Analysis on coarse sampling ratio.} 
$S$ controls the number of coarse sampled frames, determining the degree of redundancy removal. As shown in \cref{fig:variation} (b), the highest coarse sampling rate ($S=0.50$) fails to yield the best results, even though it provides more candidates for fine sampling. This indicates that redundancy elimination effectively narrows the search scope, enhancing question relevant retrieval.

\noindent\textbf{Analysis on pruning ratio.} 
\cref{fig:variation} (c) shows that full ZAP consistently outperforms ZAP\textsubscript{phy} (using physics-aware pruning only), especially at high pruning ratios. Interestingly, ZAP with low pruning ($K=0.2$) significantly outperforms the original model by 2 points. This suggests that proper pruning can filter out noisy signals and enhance the understanding and reasoning capabilities of LLM.

\begin{table}[t]
    \centering
    \caption{\textbf{Comparison with alternative sampling strategies.} Please refer to \cref{sec:ablation_study} for the explanations of these abbreviations.}
    \setlength{\tabcolsep}{9pt}
    \renewcommand{\arraystretch}{1}
    \resizebox{1.0\linewidth}{!}{
      \begin{tabular}{c|c|cc}
        \toprule
        \multicolumn{2}{c|}{Method} & EventBench$\downarrow$ & LongVideoBench$\downarrow$ \\
        
        \midrule
        \multirow{3}{*}{\shortstack{Coarse\\Sampling}} & UNI & 53.09 & 59.54 \\
        & TOP & 54.09 & 58.94 \\
        & CS & \textbf{55.09} & \textbf{60.43} \\
        
        \midrule
        
        \multirow{2}{*}{\shortstack{Fine\\Sampling}} & TOP & 53.89 & 60.21 \\
        & BIN & \textbf{55.09} & \textbf{60.43} \\
        
        \bottomrule
      \end{tabular}
    }
    \label{tab:CS}
\end{table}

\begin{table}[t]
    \centering
    \caption{\textbf{Ablation study on ZAP.} ``Sem.'' denotes semantic-aware pruning, ``Phy.'' denotes physics-aware pruning, and ``Ada.'' denotes adaptive pruning ratio.}
    \setlength{\tabcolsep}{6pt}
    \renewcommand{\arraystretch}{1}
    \resizebox{1.0\linewidth}{!}{
      \begin{tabular}{ccc|cc}
        \toprule
        Sem. & Phy. & Ada. & EventBench$\downarrow$ & LongVideoBench$\downarrow$ \\
        \midrule
        \checkmark &   &   & 52.10 & 58.12 \\
          & \checkmark &   & 53.29 & 58.64 \\
        \checkmark & \checkmark &   & 53.89 & 59.24 \\
        \checkmark & \checkmark & \checkmark & \textbf{55.49} & \textbf{60.66} \\
        \bottomrule
      \end{tabular}
    }
    \label{tab:ablation_ZAP}
\end{table}

\section{Conclusion}
This work proposes a training-free, event-guided framework to address the efficiency challenge in Video-LLMs. It is built on a dual exploitation of event data: a coarse-to-fine sampler that leverages the change-triggered property of events to eliminate temporal redundancy, and an adaptive token pruner that utilizes their inherent visual saliency to filter uninformative tokens. We validate our method on EventBench, our newly constructed benchmark with real events, as well as on three general video benchmarks with simulated events. The comprehensive empirical evaluation across all four benchmarks confirms the effectiveness of our event-guided strategy, opening a new avenue for event-assisted LLMs.

{
    \small
    \bibliographystyle{ieeenat_fullname}
    \bibliography{main}
}


\end{document}